\documentclass[10pt,twocolumn,letterpaper]{article}

\usepackage{iccv}
\usepackage{times}
\usepackage{epsfig}
\usepackage{graphicx}
\usepackage{amsmath}
\usepackage{amssymb}
\usepackage{subfig}
\usepackage{array}
\usepackage[breaklinks=true,bookmarks=false]{hyperref}

\iccvfinalcopy % *** Uncomment this line for the final submission

\def\iccvPaperID{****} % *** Enter the ICCV Paper ID here

\ificcvfinal\fi

\begin{document}

%%%%%%%%% TITLE
\title{Generating High-Resolution Fashion Model Images Wearing Custom Outfits}

\author{G\"{o}khan Yildirim\qquad Nikolay Jetchev\qquad Roland Vollgraf\qquad Urs Bergmann\\
Zalando Research\\
{\tt\small \{gokhan.yildirim,nikolay.jetchev,roland.vollgraf,urs.bergmann\}@zalando.de}
}

\maketitle
\ificcvfinal\thispagestyle{empty}\fi

%%%%%%%%% ABSTRACT
\begin{abstract}
   Visualizing an outfit is an essential part of shopping for clothes. Due to the combinatorial aspect of combining fashion articles, the available images are limited to a pre-determined set of outfits. In this paper, we broaden these visualizations by generating high-resolution images of fashion models wearing a custom outfit under an input body pose. We show that our approach can not only transfer the style and the pose of one generated outfit to another, but also create realistic images of human bodies and garments.
\end{abstract}

%%%%%%%%% BODY TEXT
\vspace{-0.6cm}
\section{Introduction}
Fashion e-commerce platforms simplify apparel shopping through search and personalization. A feature that can further enhance user experience is to visualize an outfit on a human body. Previous studies focus on replacing a garment on an already existing image of a fashion model~\cite{cagan, viton} or on generating low-resolution images from scratch by using pose and garment color as input conditions~\cite{clothing-people}. In this paper, we concentrate on generating high-resolution images of fashion models wearing desired outfits and given poses.

In recent years, advances in Generative Adversarial Networks (GANs)~\cite{gan} enabled sampling realistic images via implicit generative modeling. One of these improvements is Style GAN~\cite{stylegan}, which builds on the idea of generating high-resoluton images using Progressive GAN~\cite{progressive-gan} by modifying it with Adaptive Instance Normalization (AdaIN)~\cite{adain}. In this paper, we employ and modify Style GAN on a dataset of model-outfit-pose images under two settings: We first train the vanilla Style GAN on a set of fashion model images and show that we can transfer the outfit color and body pose of one generated fashion model to another. Second, we modify Style GAN to condition the generation process on an outfit and a human pose. This enables us to rapidly visualize custom outfits under different body poses and types.

\section{Outfit Dataset}
\begin{figure}[h]
    \centering
    \includegraphics[width=0.46\textwidth]{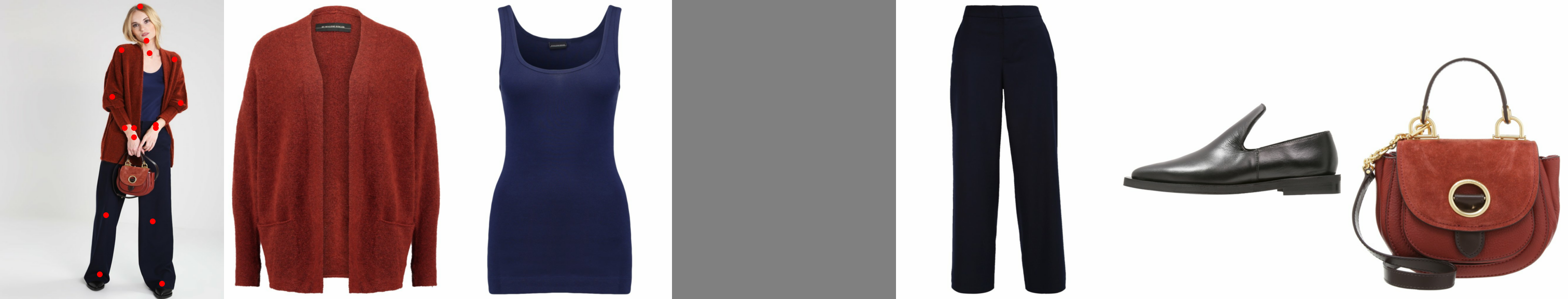}\\
    \includegraphics[width=0.46\textwidth]{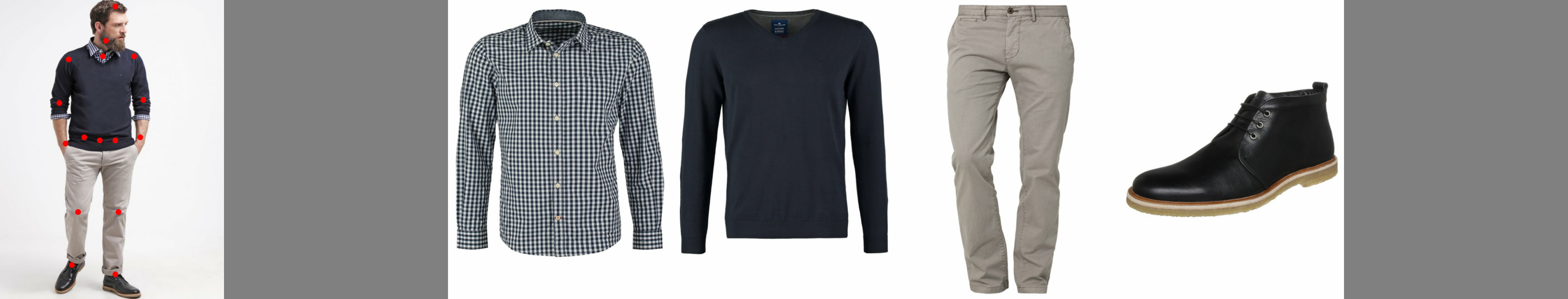}
    \caption{Samples from our dataset (red markers represent the extracted keypoints).}
    \label{fig:dataset}
\end{figure}

We use a proprietary image dataset with around 380K entries. Each entry in our dataset consists of a fashion model wearing an outfit with a certain body pose. An outfit is composed of a set of maximum 6 articles. In order to obtain the body pose, we extract 16 keypoints using a deep pose estimator~\cite{pose-extract}. In Figure~\ref{fig:dataset}, we visualize a few samples from our dataset. The red markers on the fashion models represent the extracted keypoints. Both model and articles images have a resolution of $1024 \times 768$ pixels.

\vspace{-0.4cm}
\section{Experiments}
\begin{figure}[!h]
    \centering
    \subfloat[Unconditional]{
    \includegraphics[trim=40 360 80 0 clip,width=0.91\columnwidth]{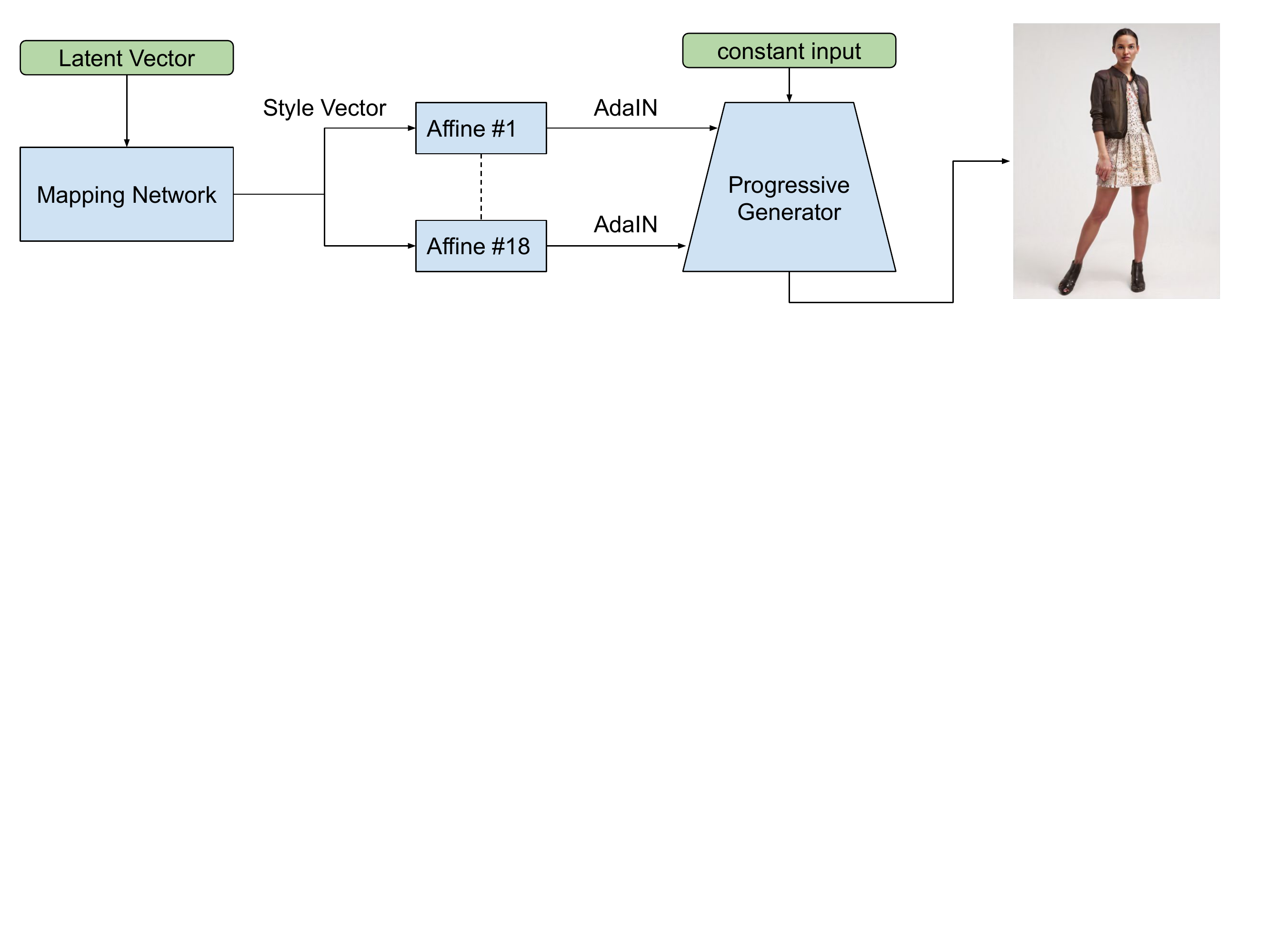}\label{uncond}}\\
    \subfloat[Conditional]{
    \includegraphics[trim=40 200 80 0 clip,width=0.91\columnwidth]{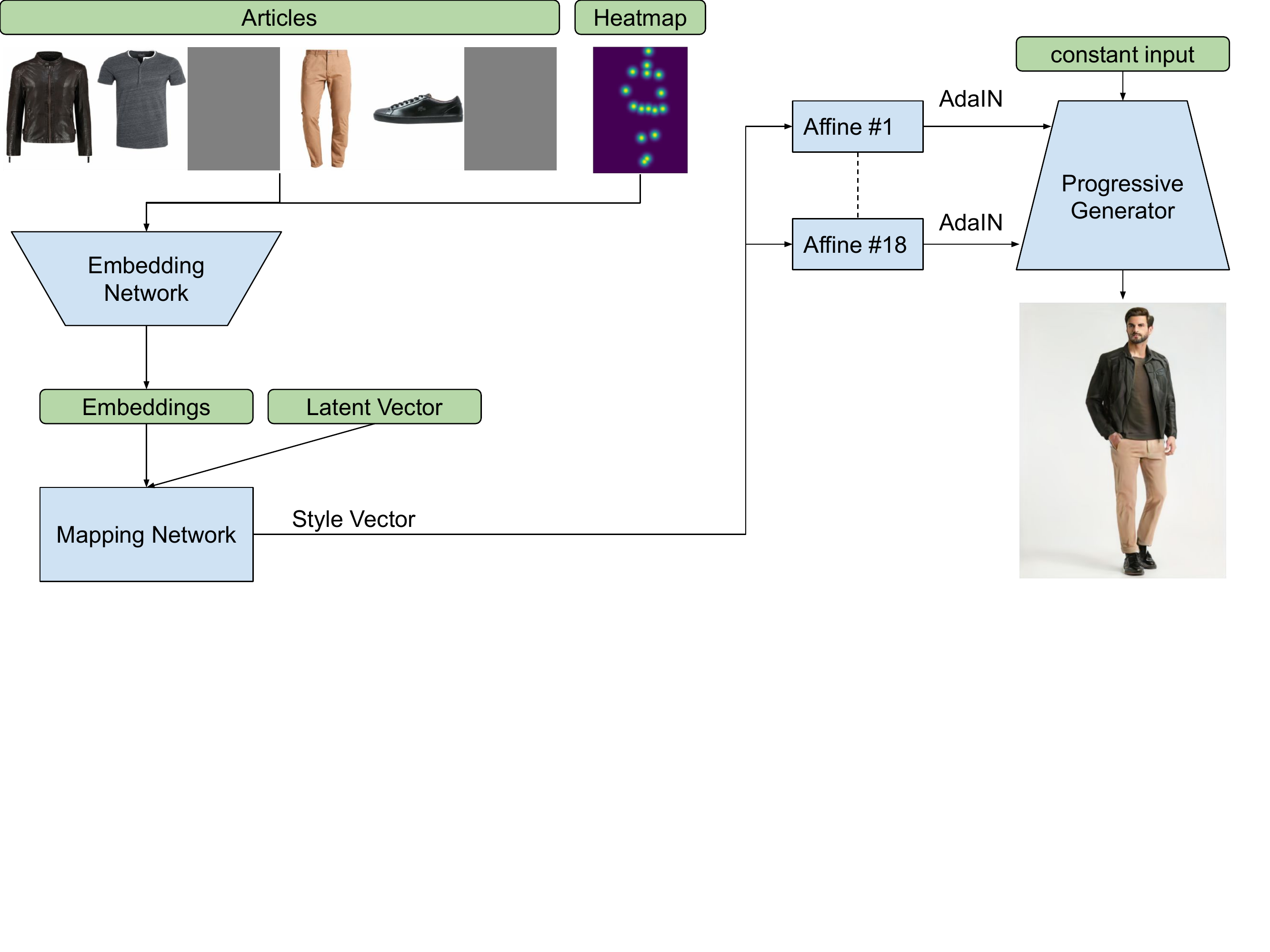}\label{cond}}
    \caption{The flowcharts of our (a) unconditional and (b) conditional GANs.}
    \label{fig:flowchart}
\end{figure}

The flowchart for the unconditional version of Style GAN is illustrated in Figure~\ref{fig:flowchart}\subref{uncond}. We have 18 generator layers that receive an affinely transformed copy of the style vector for adaptive instance normalization. The disciminator is identical to the original Style GAN. We train this network for around four weeks on four NVIDIA V100 GPUs, resulting in 160 epochs.

In the conditional version, we modify Style GAN with an embedding network as shown in Figure~\ref{fig:flowchart}\subref{cond}. Inputs to this network are the six article images (in total 18 channels) and a 16-channel heatmap image that is computed from 16 keypoints. The article images are concatenated with fixed ordering for semantic consistency across outfits. We can see this ordering in Figure~\ref{fig:dataset}. If an outfit does not have an article on a particular semantic category, it is filled with an empty gray image. The embedding network creates a 512-dimensional vector, which is concatenated with the latent vector in order to produce the style vector. This model is also trained for four weeks (resulting in 115 epochs). The discriminator in the conditional model uses a separate network to compute the embedding for the input articles and heatmaps, which is then used to compute a final score using~\cite{gan-converge}.
\vspace{-0.2cm}

\vspace{-0.3cm}
\subsection{Unconditional}
In Figure~\ref{fig:unconditional}, we illustrate images that are generated by the unconditional model. As we can see, not only the articles, but also the human body parts are realistically generated at the maximum resolution of $1024 \times 768$ pixels.

\begin{figure*}[!h]
    \centering
    \includegraphics[width=0.92\textwidth]{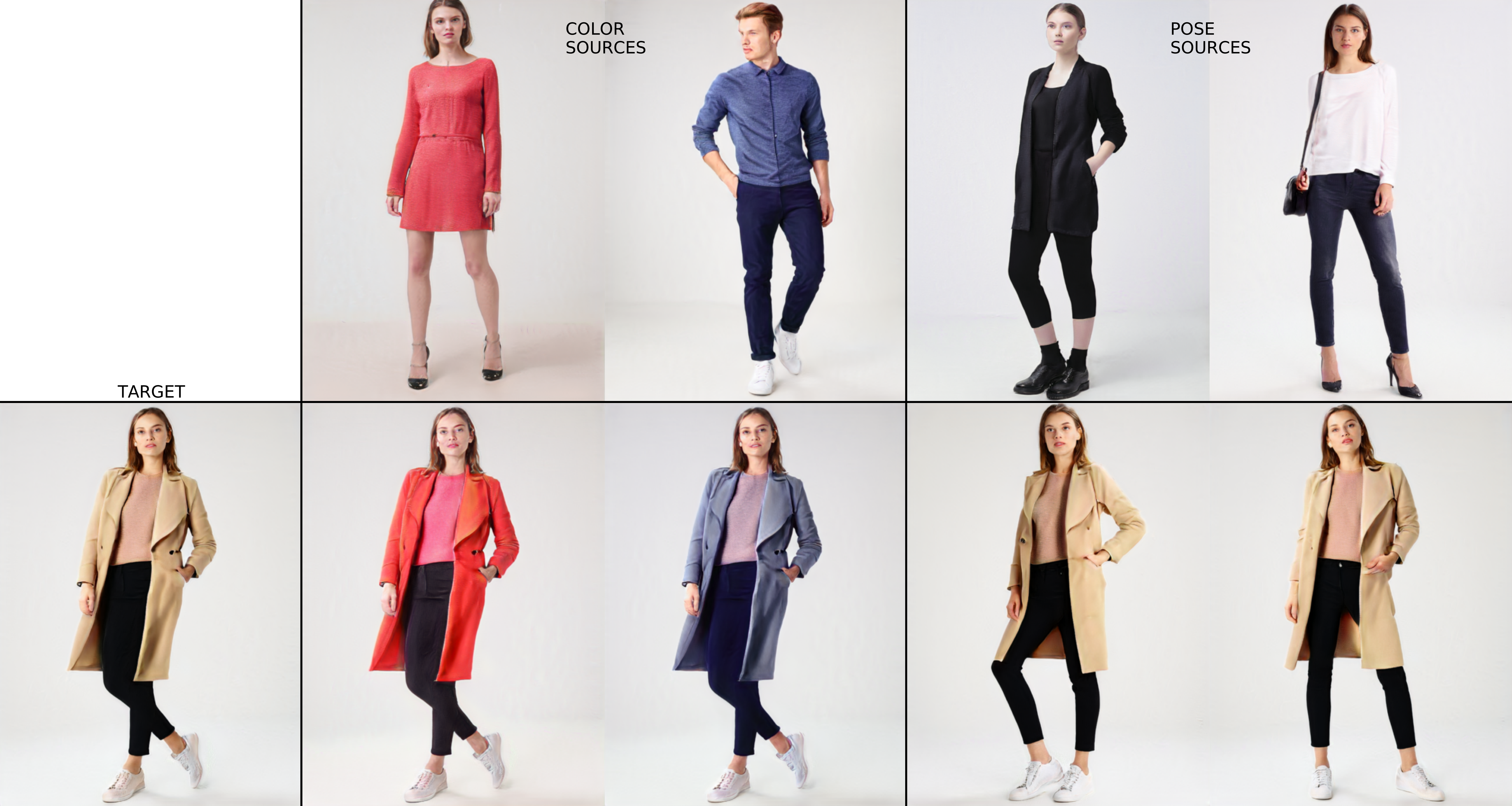}
    \caption{Transferring the colors of an outfit or a body pose to a different generated model.}
    \label{fig:color-pose-transfer}
\end{figure*}

During the training, one can regularize the generator by switching the style vectors for certain layers. This has the effect of transferring information from one generated image to another. In Figure~\ref{fig:color-pose-transfer}, we illustrate two examples of information transfer. First, we broadcast the same source style vector to layers 13 to 18 (before the affine transformations in Figure~\ref{fig:flowchart}) of the generator, which transfers the color of the source outfit to the target generated image, as shown in Figure~\ref{fig:color-pose-transfer}. If we copy the source style vector to earlier layers, this transfers the source pose. In Table~\ref{tab:layer-feeding}, we show which layers we broadcast the source and the target style vectors to achieve the desired transfer effect.

\begin{table}[h]
    \centering
    \begin{tabular}{|c|c|c|}
    \hline
       & Color Transfer & Pose Transfer\\
       \hline
       Source & 13-18 & 1-3\\
       \hline
       Target & 1-12 & 4-18\\
       \hline
    \end{tabular}
    \caption{Layers to broadcast the style vector.}
    \label{tab:layer-feeding}
\end{table}

\subsection{Conditional}
\begin{figure*}[!h]
    \centering
    \subfloat[Outfit \#1]{\includegraphics[width=0.48\textwidth]{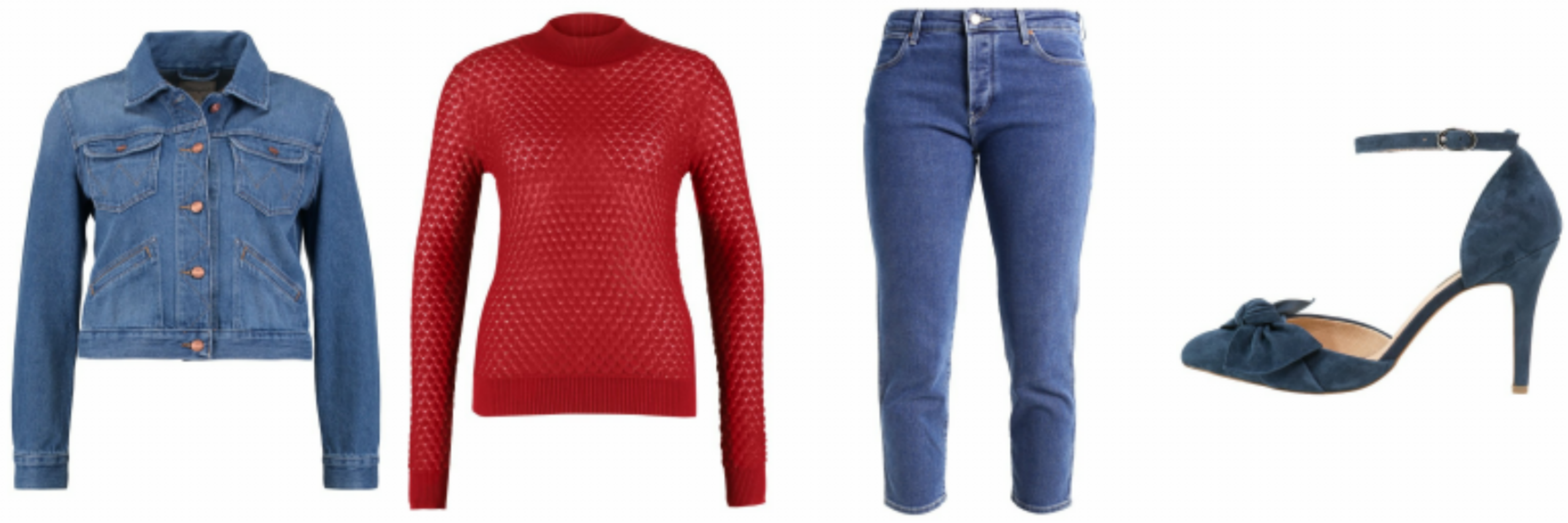}\label{outfit1}}\quad\quad\quad
    \subfloat[Outfit \#2]{\includegraphics[width=0.36\textwidth]{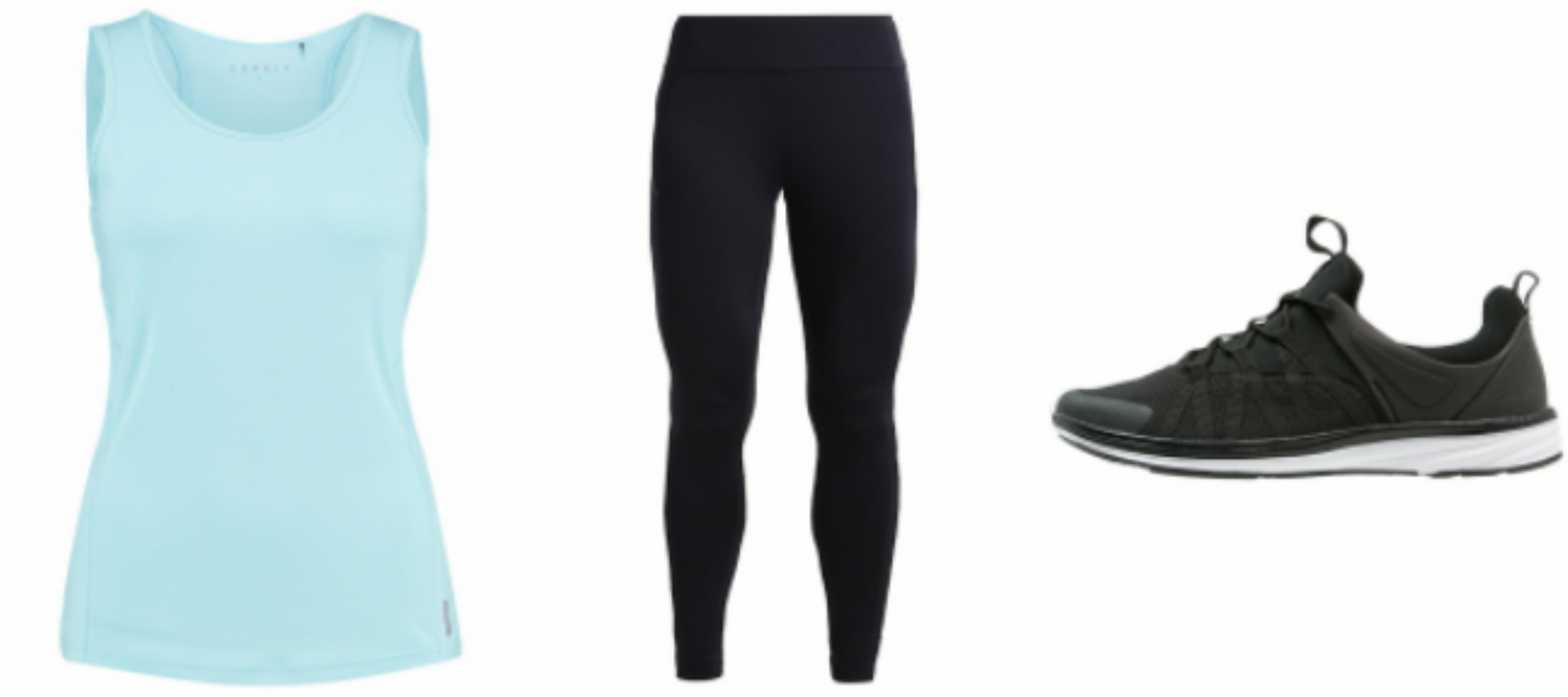}\label{outfit2}}\\
    \subfloat[Generated model images with outfit \#1]{\includegraphics[width=0.90\textwidth]{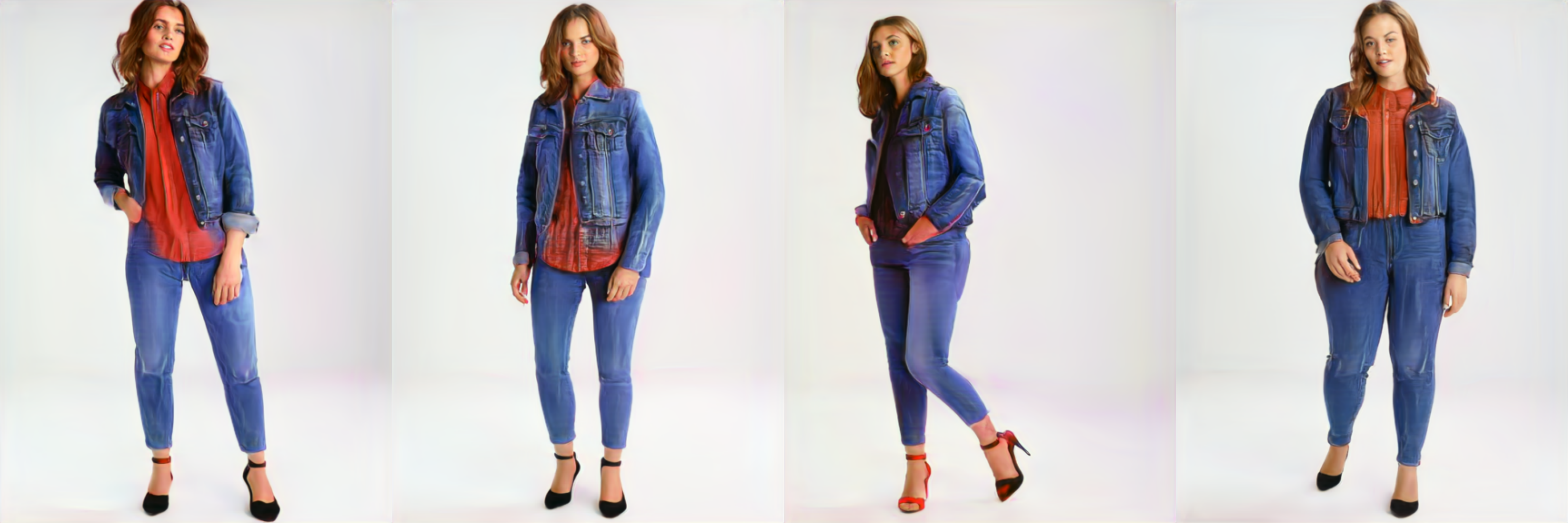}\label{condgen1}}\\
    \subfloat[Generated model images with outfit \#2]{\includegraphics[width=0.90\textwidth]{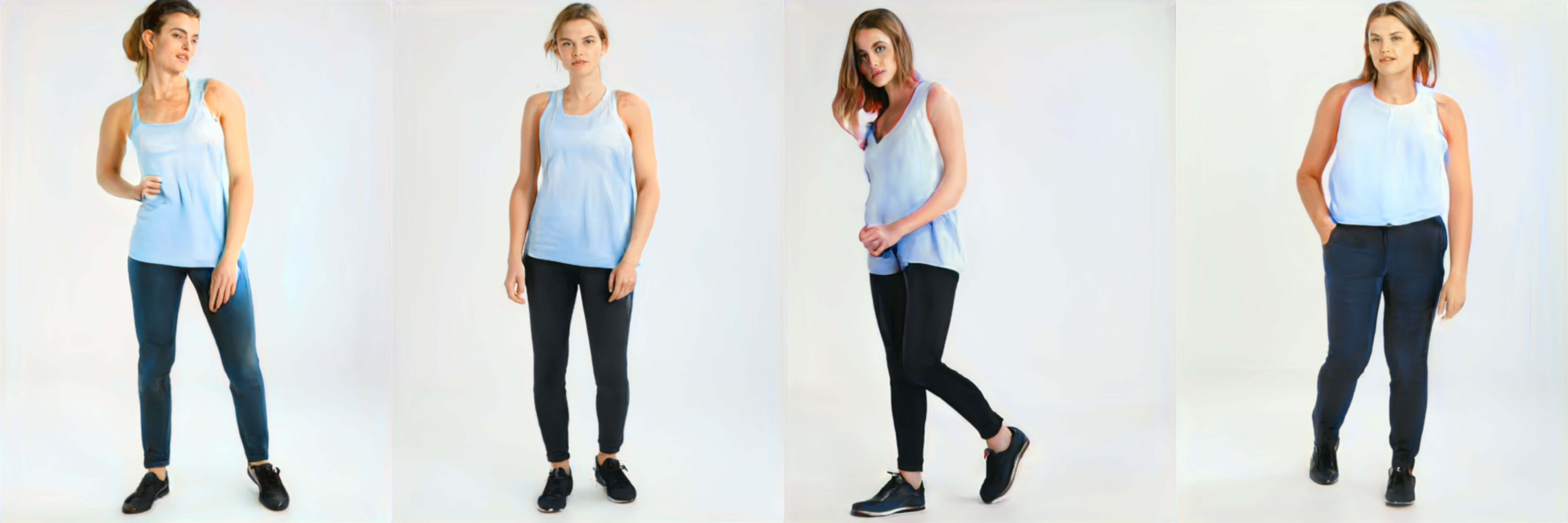}\label{condgen2}}\\
    \subfloat[Generated model images with outfit \#2 and the jacket from outfit \#1]{\includegraphics[width=0.90\textwidth]{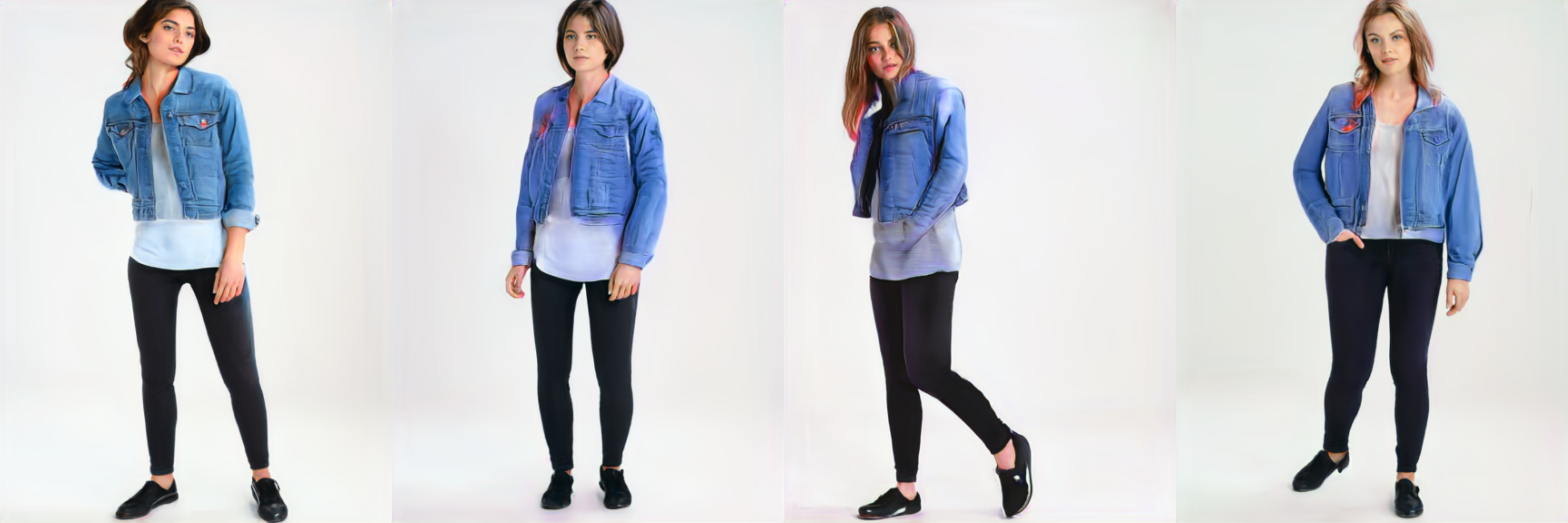}\label{condgen3}}
    \caption{Two different outfits (a) and (b) are used to generate model images in (c) and (d). (e) The jacket from outfit \#1 is added to outfit \#2 to customize the visualization.}
    \label{fig:outfit-visualization}
\end{figure*}

After training our conditional model, we can input a desired set of articles and a pose to visualize an outfit on a human body as shown in Figure~\ref{fig:outfit-visualization}. We use two different outfits in Figure~\ref{fig:outfit-visualization}\subref{outfit1} and \subref{outfit2}, and four randomly picked body poses to generate model images in Figure~\ref{fig:outfit-visualization}\subref{condgen1} and \subref{condgen2}, respectively. We can observe that the articles are correctly rendered on the generated bodies and the pose is consistent across different outfits. In Figure~\ref{fig:outfit-visualization}\subref{condgen3}, we visualize the generated images using a custom outfit by adding the jacket from the first outfit to the second one. We can see that the texture and the size of the denim jacket are correctly rendered on the fashion model. Note that, due to the spurious correlations within our dataset, the face of a generated model might vary depending on the outfit and the pose.

In our dataset, we have fashion models with various body types that depend on their gender, build, and weight. This variation is implicitly represented through the relative distances between extracted keypoints. Our conditional model is able to capture and reproduce fashion models with different body types as shown in the fourth generated images in Figure~\ref{fig:outfit-visualization}. This result is encouraging, and our method might be extended in the future to a wider range of customers through virtual try-on applications. 

\subsection{Quantitative Results}
\begin{figure*}[!t]
    \centering
    \includegraphics[width=0.225\textwidth]{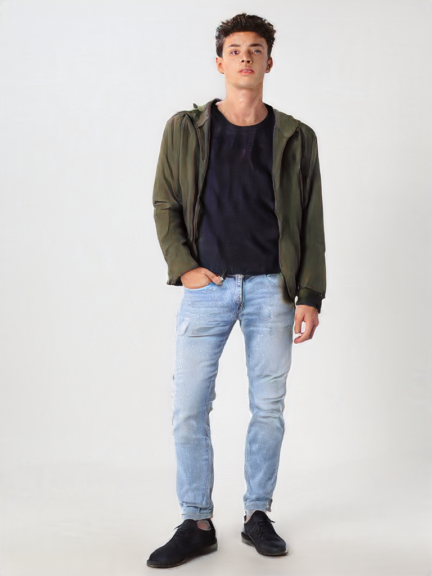}
    \includegraphics[width=0.225\textwidth]{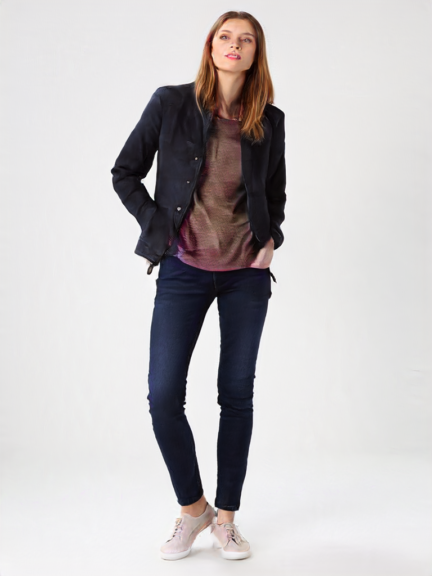}
    \includegraphics[width=0.225\textwidth]{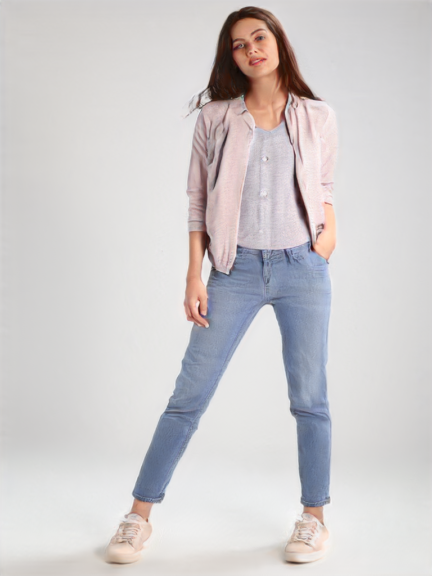}
    \includegraphics[width=0.225\textwidth]{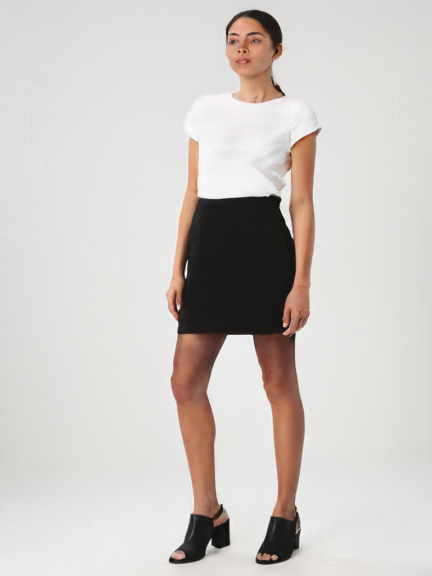}
    \caption{Model images that are generated by the unconditional Style GAN.}
    \label{fig:unconditional}
\end{figure*}

We measure the quality of the generated images by computing the Fr\'{e}chet Inception Distance (FID) score~\cite{fid-score} of the unconditional and conditional GANs. As we can see from Table~\ref{tab:fid-score}, the unconditional GAN produces higher quality images, which can be observed by comparing Figure~\ref{fig:unconditional} and Figure~\ref{fig:outfit-visualization}. The conditional discriminator has the additional task of checking whether the input outfit and pose are correctly generated. This might cause a trade-off between image quality (or `realness') and the ability to directly control the generated outfit and pose. 
\begin{table}[h]
    \centering
    \begin{tabular}{|l|c|c|}
    \hline
       & FID Score & Training Epochs\\
       \hline
       Unconditional & 5.15 & 115\\
       \hline
       Conditional & 9.63 & 115 \\
       \hline
    \end{tabular}
    \caption{FID Score for the models.}
    \label{tab:fid-score}
\end{table}
\vspace{-0.2cm}
\section{Conclusion}
In this paper, we proposed two ways to generate high-resolution images of fashion models. First, we showed that the unconditional Style GAN can be used to transfer the style/color and the pose between generated images via swapping the style vectors at specific layers. Second, we modified Style GAN with an embedding network, so that we can generate images of fashion models wearing a custom outfit with a give pose. As future work, we plan to improve the image quality and consistency of the conditional model on more challenging cases, such as generating articles with complicated textures and text.

\vspace{-0.2cm}
{\small
\bibliographystyle{ieee_fullname}
\bibliography{egbib}

\begin{thebibliography}{10}\itemsep=-1pt

\bibitem{gan}
Ian Goodfellow, Jean Pouget-Abadie, Mehdi Mirza, Bing Xu, David Warde-Farley,
  Sherjil Ozair, Aaron Courville, and Yoshua Bengio.
\newblock Generative adversarial nets.
\newblock In {\em NIPS}. 2014.

\bibitem{viton}
Xintong Han, Zuxuan Wu, Zhe Wu, Ruichi Yu, and Larry~S. Davis.
\newblock Viton: An image-based virtual try-on network.
\newblock {\em CVPR}, 2017.

\bibitem{fid-score}
Martin Heusel, Hubert Ramsauer, Thomas Unterthiner, Bernhard Nessler, and Sepp
  Hochreiter.
\newblock Gans trained by a two time-scale update rule converge to a local nash
  equilibrium.
\newblock In {\em NIPS}. 2017.

\bibitem{adain}
Xun Huang and Serge~J. Belongie.
\newblock Arbitrary style transfer in real-time with adaptive instance
  normalization.
\newblock {\em ICCV}, 2017.

\bibitem{cagan}
Nikolay Jetchev and Urs Bergmann.
\newblock The conditional analogy gan: Swapping fashion articles on people
  images.
\newblock In {\em ICCV Workshops}, 2017.

\bibitem{progressive-gan}
Tero Karras, Timo Aila, Samuli Laine, and Jaakko Lehtinen.
\newblock Progressive growing of gans for improved quality, stability, and
  variation.
\newblock {\em ICLR}, 2017.

\bibitem{stylegan}
Tero Karras, Samuli Laine, and Timo Aila.
\newblock A style-based generator architecture for generative adversarial
  networks.
\newblock {\em CVPR}, 2019.

\bibitem{clothing-people}
Christoph Lassner, Gerard Pons-Moll, and Peter~V. Gehler.
\newblock A generative model of people in clothing.
\newblock In {\em ICCV}, 2017.

\bibitem{gan-converge}
Lars Mescheder, Andreas Geiger, and Sebastian Nowozin.
\newblock Which training methods for gans do actually converge?
\newblock In {\em ICML}, 2018.

\bibitem{pose-extract}
Ke Sun, Bin Xiao, Dong Liu, and Jingdong Wang.
\newblock Deep high-resolution representation learning for human pose
  estimation.
\newblock In {\em CVPR}, 2019.

\end{thebibliography}
}

\end{document}